\documentclass{article}
\usepackage{ijcai19}

\usepackage{times}
\usepackage{soul}
\usepackage{url}
\usepackage[hidelinks]{hyperref}
\usepackage[utf8]{inputenc}
\usepackage[small]{caption}
\usepackage{graphicx}
\usepackage{amsmath}
\usepackage{booktabs}
\usepackage{algorithm}
\usepackage{algorithmic}
\usepackage{xcolor}
\usepackage{colortbl}

\renewcommand{\arraystretch}{1.5}
\usepackage{supertabular}

\title{Inception-inspired LSTM for Next-frame Video Prediction}

\author{
Matin Hosseini\and Anthony S. Maida \and Majid Hosseini
\and Gottumukkala Raju \\
\affiliations
School of Computing and Informatics, University of Louisiana at Lafayette
\emails
\{mxh0212,maida,seyedmajid.hosseini1,raju\}@louisiana.edu
}

\begin{document}

\maketitle

\begin{abstract}

The problem of video frame prediction has received much interest due to its relevance to many computer vision applications such as autonomous vehicles or robotics. Supervised methods for video frame prediction rely on labeled data, which may not always be available. In this paper, we provide a novel self-supervised deep-learning method called Inception-based LSTM for video frame prediction. The general idea of inception networks is to implement wider networks instead of deeper networks. This network design was shown to improve the performance of image classification. The proposed method is evaluated on both Inception-v1 and Inception-v2 structures.  The proposed Inception LSTM methods are compared with convolutional LSTM when applied using PredNet predictive coding framework for both the KITTI and KTH data sets. We observed that the Inception based LSTM outperforms the convolutional LSTM. Also, Inception LSTM has better prediction performance compared to Inception v2 LSTM. However, Inception v2 LSTM has a lower computational cost compared to Inception LSTM.

\vspace{12pt}
\end{abstract}

\section{Introduction}

Video frame prediction has received much interest in computer vision and deep learning due to its application to several domains such as video transcoding \cite{hosseini2017enabling}, anomaly detection \cite{anomaly}, autonomous car guidance \cite{xu2017end}, and robotic motion planning \cite{finn2017deep}. These applications use video frame predictions to understand the dynamics of the environment. The human brain has the ability to forecast anticipated changes through sensory perception and intuition. For example, a goalkeeper in a soccer game can anticipate the ball movement direction from the position of the players and previous ball movement. The theory of predictive coding illustrates the interaction of feed-forward and feed-backward information flow in the brain hierarchy \cite{Friston2005}. 
Inspired by the theory of predictive coding, Lotter et al. \cite{lotter2016} proposed an LSTM-based architecture for frame prediction. The Lotter model improved the representation of the next frame using a convolutional LSTM by reducing the error in frame prediction. 
In contrast to a standard LSTM, a convolutional LSTM processes a multi-channel image and applies convolutions to those images instead of fully connected.

A standard convolutional LSTM uses a single size of kernel for each of its gates.  This paper proposes an Inception-inspired LSTM that uses multiple-kernels of different sizes inside the structure of LSTM for each input gate.
The outputs of each kernel are then concatenated as a stack of images and passed to the gate activation.\\
We designed an Inception LSTM for video prediction using two different Inception versions. The inception LSTM replaces convolution with an Inception module to employ benefits of manipulating different kernels that can capture different motions.

This paper is organized as follows. 
Section~2 reviews some recent innovations in video prediction. 
In section~3 we talk about Inception LSTM in detail. 
In sections 4 and 5 we discuss evaluation and results, and finally we conclude in section 6.

%%%%%%%%%%%%%%%%%%%%%%%%% 

\section{Related work}

%Deep learning has recently made significant progress in supervised learning. 
%For example, 
Convolutional neural networks (CNNs) are successful in image classification and object 
detection as seen the result of ALEXNET \cite{alexnet}, VGG \cite{vgg}, ResNet\cite{resnet} and GoogLeNet \cite{googlenet}.
GoogLeNet uses Inception CNN layers in its architectures which is considered a major breakthrough in classification. The idea behind the inception network is to have a wider network instead of a deeper one. The Inception neural network offers better predictive performance without increase in computational cost. 
An inception layer uses a different kernel size for each convolution that has a different 
receptive field. Here, the network starts with the smallest kernel of size 1x1 to a kernel of size 5x5. This makes the network adaptable for different scales of objects \cite{gitbook}.  
The results of different kernels are stacked and  passed on to the consecutive layers. 

Various versions of Inception have been proposed in the literature \cite{szegedy2016rethinking}. Inception Version 2 proposed stacking two 3x3 kernels in place of using one 5x5 that reduces representational bottleneck Although this may seem counter intuitive, a 5x5 convolution is 2.78 times more expensive than a 3x3 convolution \cite{szegedy2016rethinking}.

Prediction of future frames in video content is difficult using non-recurrent neural networks  \cite{Zhou2016} because the information retained about the sequence of frames does not accumulate over time. These methods have limited capability and are not suitable for complex tasks \cite{Selva2018}. Recently several architectures have been proposed using recurrent neural networks such as LSTM or convolutional LSTM for video prediction.
Villegas et al. \cite{motion} used different frames to decompose the motion and content. 
Chelsea and Goodfellow \cite{IEEEhowto:chelsea} developed a video prediction method that modeled pixel prediction.
This model can generalize to unseen objects because they predict motion.
Viorica \cite{opticalflow} used an optical flow based architecture for feature extraction in an unsupervised manner.   

Lotter et al. \cite{Lotter2015} proposed a predictive generative network for predicting video frames. 
Their first model consisted of three different components: A convolution neural network, de-convolution, and LSTM component. 

Shi \cite{shi2015} proposed a convolutional LSTM that can be used for video prediction. Our Inception LSTM is inspired by convolutional LSTM.
 
PredNet \cite{lotter2016} is a convolutional LSTM-based predictive coding model, implemented within multi-layer network. The lowest layer predicts the next frame in a video sequence and is also guided by top-down context supplied by the higher layers. 
The model propagates the error laterally using recurrent connections and vertically which is inspired by the Rao and Ballard model \cite{Rao} which is a seminal study in predictive coding.

%TODO Matin check shi 
The formula proposed by Shi et al. \cite{shi2015} uses peephole connections which were removed in Lotter et al. implementation. 
We followed the Lotter architecture \cite{lotter2016}
known as PredNet for video prediction in our Inception LSTM model. 

We proposed an inception LSTM that benefits from the inception method using three different kernel sizes at the same time in each layer structure to have a better compatibility with all object scales in video frames.

%%%%%%%%%%%%%%%%%%%%%%%%% Begin figure
\begin{figure}[ht]
\centering
\includegraphics[scale=0.25]{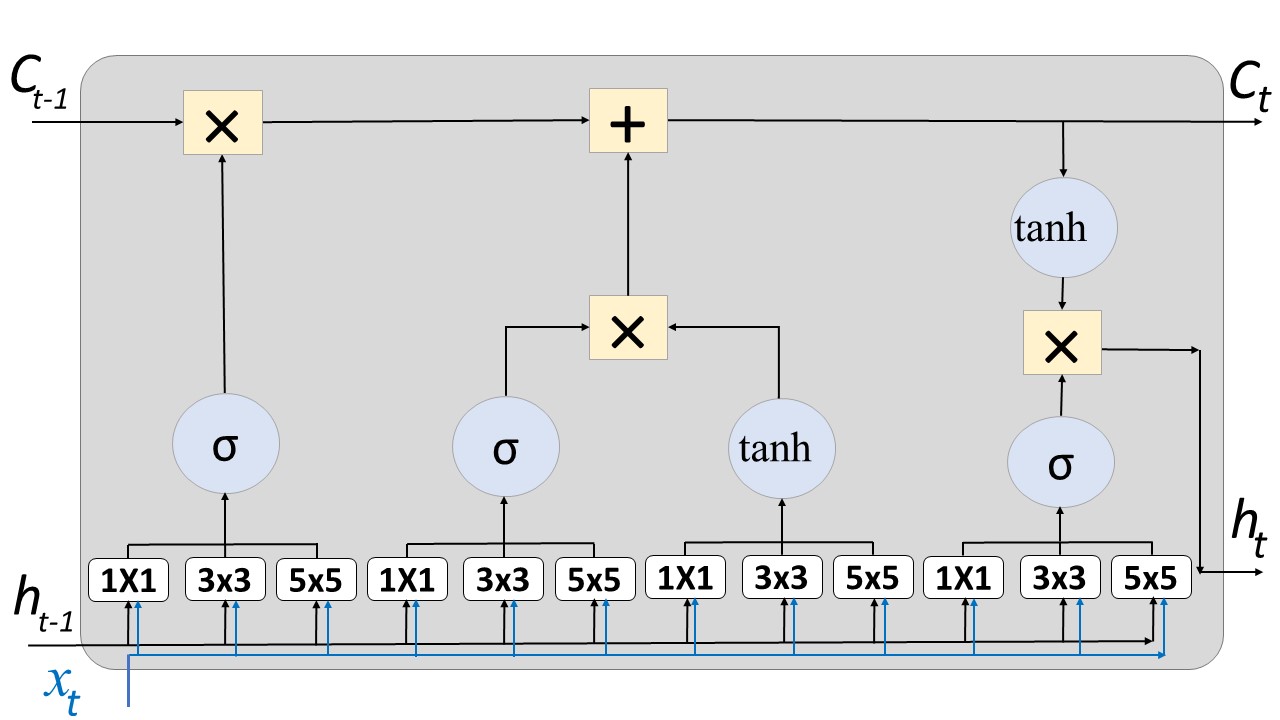}
\caption{Our Inception-inspired Version 1 LSTM module.
Each of the gates, as well as the input update node, receives the stacked outputs of
three convolutions.
Version 2 replaces the 5x5 convolutions with a sequence of 3x3 convolutions for parameter reduction.}
\label{fig:InceptionLSTM}
\end{figure}
%%%%%%%%%%%%%%%%%%%%%%%%%

\section{The Inception-Inspired LSTM}
LSTMs are among the state-of-the-art at learning long-term dependencies\cite{wojtkiewicz2019hour} that show up in different predictive sequence models. An LSTM has an internal cell state \cite{Greff2017} that is modified across time steps in an additive fashion.
The standard LSTM also has three gates (input, forget, and output) as well as peephole connections. The input gate plays the most important role in an LSTM\@ by learning how valuable the current input is for further predictions. 
The $h$ state is the component that helps short-term memories. The forget gate has a significant role in performance and learns to determine whether previous history is no longer relevant to future predictions. The output gate is the least important component.The forget gate and activation function are the most important part in most of the designs \cite{Greff2017}.
%%%%%%%%%%%%%%%%%%%%%%%%%% Begin figure
\begin{figure}[ht]
\centering
\includegraphics[scale=0.25]{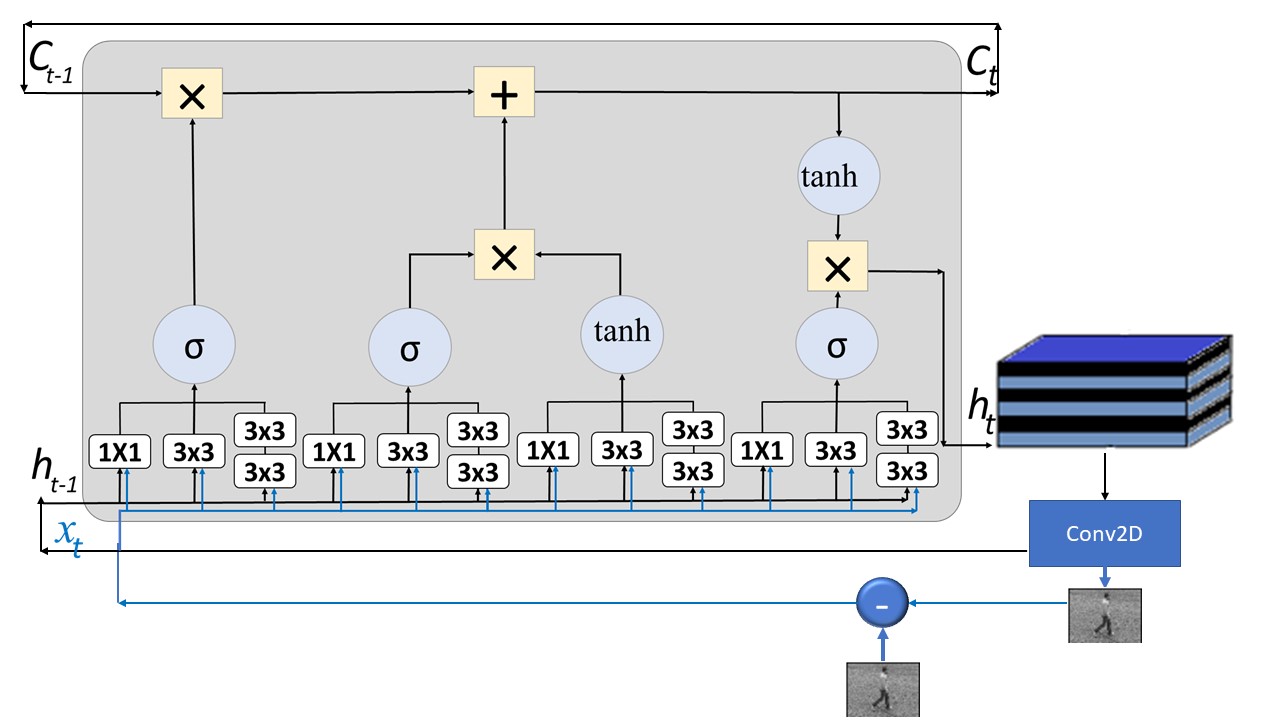}
\caption{Lowest layer of the PredNet architecture using an Inception-inspired Version 2 LSTM\@.
The inputs to the subtraction operation are the next video frame and the predicted
next video frame supplied by the LSTM\@. The  Conv2D module reduces
the channel size of the output from the Inception LSTM so that its dimensions are compatible with the target frame for video subtraction.}
\label{fig:architecture}
\end{figure}
%%%%%%%%%%%%%%%%%%%%%%%%%

Other architectures have been proposed,such as the GRU \cite{Greff2017}, to perform the same function as an LSTM cell with fewer components. 
Although, there are many LSTM models, Greff et al. \cite{Greff2017} believe the standard LSTM is overall still the most effective. 
Currently,  there is no proof that the GRU has the advantage to LSTM\@. 
Specifically, Rafael et al. \cite{Jozefowicz2015} showed the LSTM and GRU performed the same by adding one bias to the 
forget gate.

A convolutional LSTM \cite{shi2015} uses convolution on the input images instead of fully connected LSTM. The convolutional LSTM uses a stack of images as its internal data structure.
It has been shown to be effective in some forms of spatio-temporal prediction.

The models using the convLSTM are dependent on the kernel size and number of layers. The formula of convolutional LSTM modifies the standard LSTM by replacing dot product operations with convoloution as follows:

\begin{subequations}
\begin{align}
i_t&=\sigma(W_{ix}*x_t+W_{ih}*h_{t-1}+b_i)\\
f_t&=\sigma(W_{fx}*x_t+W_{fh}*h_{t-1}+b_f)\\
c_t^\prime&=i_t\odot \tanh(W_{cx}*x_t+W_{ch}*h_{t-1}+b_c)\\
c_t&=f_t\odot c_{t-1}+ c_t^\prime\\
%c_t&=f_t\odot c_{t-1}+i_t\odot \tanh(W_{xc}*x_t+W_{hc}*h_{t-1}+b_c)\\
o_t&=\sigma(W_{ox}*x_t+W_{oh}*h_{t-1}+b_o)\\
h_t&=o_t\odot \tanh(c_t)
\end{align}
\end{subequations}
Each gate contains the convolution operation.The above equations describe the basic convolutional LSTM used in lotter\cite{lotter2016}
We propose an Inception-inspired LSTM that has the advantage of allowing convolution with different kernel sizes. 
In this part, we review two versions of Inception LSTM based on the configuration of Inception network. Figure \ref{fig:InceptionLSTM} shows a version 1 Inception net embedded within an LSTM cell.

Below we give the equations used in Inception LSTM version 1. The inputs to all gates are the stacking of convolution operations with different kernel sizes.  Based on Shi's study \cite{shi2015} different kernels can capture different motion magnitudes. The equations are as follows:
\begin{subequations}
\begin{align}
i_t&=\sigma
\left[
  \begin{array}{c}
  W_{i1\times1}*[x_t,h_{t-1}], \\
  W_{i3\times3}*[x_t,h_{t-1}], \\
  W_{i5\times5}*[x_t,h_{t-1}] 
  \end{array}
\right]\\
f_t&=\sigma
\left[ \begin{array}{c}
  W_{f1\times1}*[x_t,h_{t-1}], \\
  W_{f3\times3}*[x_t,h_{t-1}], \\
  W_{f5\times5}*[x_t,h_{t-1}] 
  \end{array}
\right]\\
g_t&=\sigma
\left[ \begin{array}{c}
  W_{g1\times1}*[x_t,h_{t-1}], \\
  W_{g3\times3}*[x_t,h_{t-1}], \\
  W_{g5\times5}*[x_t,h_{t-1}] 
  \end{array}
\right]\\
o_t&=\sigma
\left[ \begin{array}{c}
  W_{o1\times1}*[x_t,h_{t-1}], \\
  W_{o3\times3}*[x_t,h_{t-1}], \\
  W_{o5\times5}*[x_t,h_{t-1}] 
  \end{array}
\right]\\
c_t&=f_t\odot c_{t-1}+i_t\odot g_t\\
h_t&=o_t\odot tanh(c_t)
\end{align}
\end{subequations}
We use the Inception design for each of the gates instead of convolution. 
Each gate has three different kernels, with sizes 1x1, 3x3, and 5x5. 
$W_{i1\times1}$ denotes the weights for the input gate and $1\times 1$ shows the kernel size. 
The output of the three convolutions (indicated by square brackets) are stacked and passed to the input gate. 
The cell state and recurrent connection ($h$) is defined similarly to the original
convolution LSTM.  
Figure \ref{fig:InceptionLSTM} shows the Inception LSTM cell.
The previous paragraph is based on Inception Version 1.
We also studied a model based on Inception Version 2 that uses fewer parameters. 
The difference between the two methods relates to the 5x5 kernel in Version 1.
In Version 2, it is replaced by a stack of two 3x3 kernels.
The equation used in Inception LSTM Version 2 for the input gate is as follows:  
\begin{subequations}
\begin{align}
i_t&=\sigma\left[ \begin{array}{cc}
  W_{i1\times1}*[x_t,h_{t-1}], \\
  W_{1i3\times3}*[x_t,h_{t-1}], \\
  W_{2i3\times3}*[W_{3i3\times3}*[x_t,h_{t-1}]] 
  \end{array}
\right ]
\end{align}
\end{subequations}

%Inception Version 2 uses two 3x3 convolution kernels instead of one 5x5. 
Analogous changes are made for the other gates.
Figure \ref{fig:architecture} shows the Inception Version 2 model
embedded in one layer of the PredNet architecture. %majid 

Figure  \ref{fig:architecture} shows  Inception LSTM v2 to handle the number of channel of input and h we used Conv2D. \color{black}  Inception Version 2 reduces the number of training parameters in comparison to Version 1.  
Version 1 has a total of
$(1+9+25)\cdot \mathsf{nc}$ training parameters for each gate in each layer. 
Inception Version 2 has $(1+9*3)\cdot \mathsf{nc}$ parameters reducing the number of training parameters by $7\cdot\mathsf{nc}$ in each gate. $\mathsf{nc}$ is a constant denoting the number
of channels in both models. The number of parameters in one layer of Inception LSTM version 1 is 6,595 whereas the ConvLSTM uses 1,081 parameters.
%%%%%%new table%%%%%%%%%%%%%%%%%%%%%%%%%%%%%%%%%%%%%%%%%%%%%%%%%%
\setlength{\arrayrulewidth}{0.3mm}
\setlength{\tabcolsep}{16pt}
\renewcommand{\arraystretch}{1.1}
\begin{center}
\begin{table*}
\begin{tabular}{ m{3.5cm}m{1cm}m{1cm}m{1cm}m{1cm}m{1cm}m{1cm}} 
\hline
 & \centering & KITTI & &  &KTH & \\
\hline
Model& MAE & MSE & SSIM& MAE & MSE & SSIM \\ [.5ex]\hline

\rowcolor[gray]{.9} ConvLSTM (2L) &0.053306 &0.010216  &0.811534 &0.044115 &0.007191 &0.867645 \\
ConvLSTM (3L) & 0.047526 & 0.008185 & 0.847114&  0.010629&0.000449&0.961777 \\ 
 
 \rowcolor[gray]{.9} ConvLSTM (4L) & 0.045806 & 0.007612 & 0.857651&0.011604&0.000573&0.955879 \\
 
 \hline
Inception v1 LSTM (2L) & 0.049663 & 0.009095& 0.833583&0.010624&0.000479&0.961024 \\ 
 
 \rowcolor[gray]{.9}Inception v1 LSTM (3L) &0.044761&0.007345 & 0.863714 &\textbf{0.010326}&\textbf{0.000406}&\textbf{0.963680} \\
 Inception v1 LSTM (4L) & \textbf{0.043640} & \textbf{0.007028}& \textbf{0.868226} & 0.010959&0.000524&0.958901\\
 \hline
 \rowcolor[gray]{.9}Inception v2 LSTM (2L) & 0.049966 & 0.009114& 0.831361&0.010752&0.000503&0.960048\\ 
 Inception v2 LSTM (3L) & 0.045021 & 0.007481&0.861877&\textbf{0.010429}&\textbf{0.000438}&\textbf{0.962630} \\
 \rowcolor[gray]{.9}Inception v2 LSTM (4L) &\textbf{0.044115}&\textbf{0.007191}&\textbf{0.867645} & 0.010637&0.000463&0.961332\\
 
 \hline

\end{tabular}
\caption{Performance on the KITTI and KTH data sets. The number in the parenthesis indicates the number of layers.  \label{table:MSEKITTI}}
\end{table*}
\end{center}
\begin{figure*}[ht]
\centering
\includegraphics[scale=1.3]{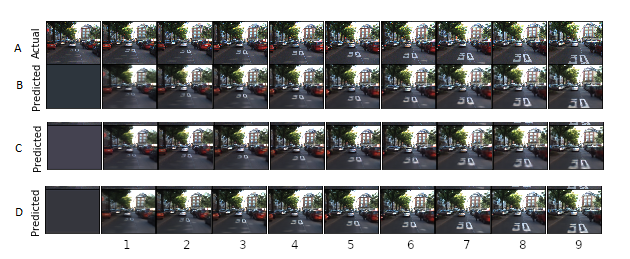}
\caption{Comparing the output of the Convolutional LSTMs and Inception LSTM on the KITTI data set. A) The actual frame. B)  Prediction using a convolutional LSTM. 
C) Prediction using Inception-inspired LSTM Version 1.
D) Prediction using Inception-inspired LSTM Version 2.}
\label{fig:result1}
\end{figure*}
\begin{figure*}[ht]
\centering
    \includegraphics[width=0.5\linewidth]{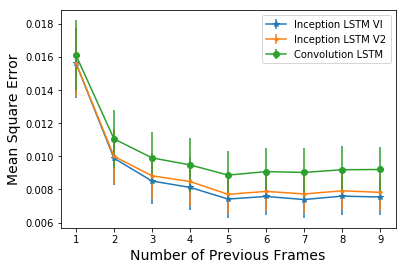}\hfil 
    \includegraphics[width=0.5\linewidth]{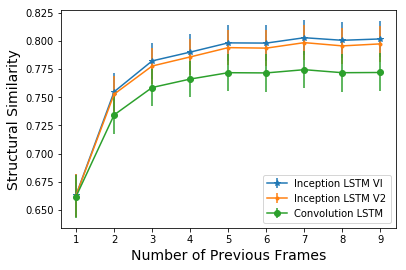}\par\medskip
\caption{KITTI data set next-frame prediction performance as a function of the number of previous frames used in the history. Left: Mean Square Error (MSE). Right: Structural Similarity (SSIM). \textbf{}}
\label{fig:result2}
\end{figure*}
%%%%%%%%%%%%%%%%%%%%%%%%%% 
\section{ Experimental setup and results}

We used inception to avoid using kernel size as a hyper parameter. In Our model there is no need to define the kernel size for inception LSTM. In this model the two variants of the Inception LSTM within the PredNet architecture \cite{lotter2016} were implemented. 
Figure \ref{fig:architecture} shows the first (lowest) layer of the PredNet architecture using 
the Inception LSTM\@. 
The higher layers repeat the same design but use different numbers of channels
for the input. We used the hard Sigmoid non-linearity instead of the original Sigmoid for gate activation  similar to \cite{lotter2016}.
For simplicity purpose, a two to four layer architecture is used for comparing our proposed model with ConvLSTM within Prednet architecture.   
We used the KITTI \cite{geiger2013vision} and KTH \cite{schuldt:04} data sets  to obtain experimental results that have RGB images.
The KITTI data set is a traffic data set that has frames from different views of the roads.
The data set is captured with a car mounted camera driving in different environments. The RGB images in KITTI data set have frame size of 160x128 pixels. The KTH contains human activity with one subject per frame. We used walking video data where each frame is a 160x120 RGB image.% majid check

The number of channels in each layer are 3, 48, 96 and 192 respectively for each of the input channel. The width and height of the input image in each layer is sub-sampled using 2x2 max pooling similar to PredNet. The source code for the implementation is made available in public at
https://github.com/matinhosseiny/Inception-inspired-LSTM-for-Video-frame-Prediction

We present the comparison of Inception LSTM with 
convolutional LSTM\@. A qualitative visual comparison of the convolutional LSTM versus Inception LSTM Versions 1 and 2 
appears in Figure \ref{fig:result1}. 
Example output using KITTI Data set is shown in Figure \ref{fig:result1} for each of the three models. Each image for rows B,C and D shows the prediction for a particular model. The column number shows the frames from previous history. The predicted images for the three models appear very similar. So three quantitative measures were used for performance comparison. The Mean Squared Error (MSE) which was the training cost function, the Structural Similarity Index (SSIM) and Mean Absolute Error.\color{black}

\begin{figure*}[h]
\centering
    \includegraphics[width=0.5\linewidth]{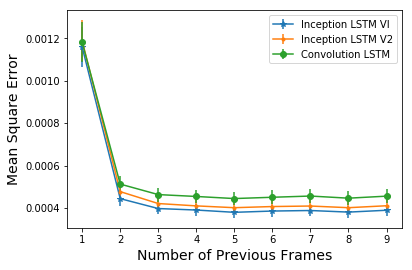}\hfil 
    \includegraphics[width=0.5\linewidth]{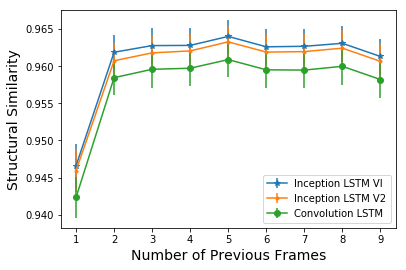}\par\medskip

\caption{KTH data set next-frame prediction performance as a function of the number of previous frames used in the history. Left: Mean Square Error (MSE). Right: Structural Similarity (SSIM).   
\textbf{}}
\label{fig:SSIM}
\end{figure*}

Figure \ref{fig:SSIM} shows SSIM and MSE performance measures on the KITTI data set. 
The sample size is 83 test videos, where nine frames per video are tested. 
We calculate the mean and 95\% confidence interval. 
Although the MSE (left) of the Inception LSTM is lower than the convolutional  LSTM,
the confidence intervals show overlap.
The SSIM measure shows the structural similarity of the predicted and ground-truth images. 
Figure \ref{fig:SSIM} (right) shows the SSIM results. 
Inception Version 1 has a better result for the SSIM measure. 
Both the MSE and SSIM results indicate that the model reaches
maximum performance after receiving at least five frames of previous history.

%Also the results in both of the diagrams shows the model gets reasonable 
%results after receiving at least five frames of data.   

% * means span columns

\begin{figure*}[ht]
\centering
\includegraphics[scale=1]{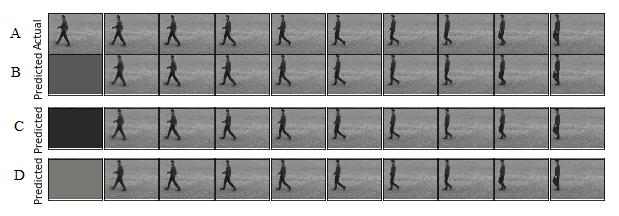}
\caption{Comparing the output of the convolutional LSTMs and Inception LSTM on the KTH data set. A) The actual frame. B)  Prediction using a convolutional LSTM. 
C) Prediction using Inception-inspired LSTM Version 1.
D) Prediction using Inception-inspired LSTM Version 2.}

\label{fig:result3}
\end{figure*}
Table \ref{table:MSEKITTI} compares the results of Inception Versions 1, 2 and 
the convolutional LSTM for the KITTI and KTH data sets. 
Inception Version I shows the best performance as judged by MSE in this experiment.
The experiment follows the exact configuration of PredNet except the number of layers is varied from two to four.
We compare the results of Inception LSTM version 1 with convolutional LSTM in PredNet architecture using KTH data set. We used the walking data set of KTH. Figure \ref{fig:result2} shows the result in comparison with ground truth.

\section{Discussion and Conclusions}

This paper proposed a novel Inception-inspired convolutional LSTM
and assessed its performance, within a predictive coding framework,
for predicting future frames in videos. 
The motivation was to see if the larger range of kernels would create
a richer feature set which would improve predictions.
We call our model Inception-inspired because it does not have all of
the characteristics of an Inception module.
Namely, max pooling is missing and the 1x1 convolution to reduce the number
of channels is shared between different kernels.
Both versions of the Inception-inspired LSTM show improved performance
over the original convolutional LSTM.

%We added the inception module to LSTM model to increase the power of the LSTM and 
%convolution LSTM. 
Possible future work is to add max pooling to the Inception-inspired LSTMs to see if that offers improvement.
On the other hand, it may be the case that the original convolutional LSTM is not the limiting factor in the PredNet predictive coding architecture and changes to the overall design may be needed.
%Inception module has the max pooling additionally the we did not consider in our model. One of the possible future work is the max pooling version. 
%%%%%%%%%%%%%%%%%%%%%%%%%%%%%%%%%%%%%%%%%%%%%%%%%%%%%%

\bibliographystyle{acm}
\bibliography{ijcai19}

\end{document}